\documentclass[11pt,a4paper]{article}
\usepackage[nohyperref]{acl2017}
\usepackage{times}
\usepackage{latexsym}

\usepackage{url}

\usepackage{amssymb}
\usepackage{amsmath}
\usepackage{amsthm}
\usepackage{enumerate}
\usepackage{graphicx}
\usepackage{algorithm}
\usepackage{algorithmic}
\usepackage{booktabs}

\aclfinalcopy 

\title{Improving a tf-idf weighted document vector embedding}

\author{Craig W. Schmidt \\
  TripAdvisor, Inc.  \\
  400 1st Avenue \\
  Needham, MA 02494 \\
  {\tt cschmidt@tripadvisor.com}}

\date{}

\begin{document}
\maketitle
\begin{abstract}

  We examine a number of methods to compute a dense vector embedding for a document in a corpus, given a set of word vectors such as those from word2vec or GloVe.  We describe two methods that can improve upon a simple weighted sum, that are optimal in the sense that they maximizes a particular weighted cosine similarity measure.

  We consider several weighting functions, including inverse document frequency (idf), smooth inverse frequency (SIF), and the sub-sampling function used in word2vec.  We find that idf works best for our applications. We also use common component removal proposed by Arora et al. as a post-process and find it is helpful in most cases. 

  We compare these embeddings variations to the doc2vec embedding on a new evaluation task using TripAdvisor reviews, and also on the CQADupStack benchmark from the literature.

\end{abstract}

\section{Introduction}

One way to represent word similarity is through a dense vector embedding, where the embedding is constructed so that similar words have vectors with a high cosine similarity. \citet{DBLP:journals/corr/abs-1301-3781,Mikolov:2013:DRW:2999792.2999959} introduced two breakthrough approaches with their skip-gram and CBOW variants of word2vec. These use neural models to predict a target word based on its context words, or vice versa. \citet{D14-1162} subsequently introduced GloVe, another successful vector embedding method that was explicitly factoring a weighted word-word co-occurrence matrix. These prediction based approaches were a significant improvement over the count based approaches that were used before \citep{baroni-dinu-kruszewski:2014:P14-1}. 

We would often like a vector embedding for a set of documents in a corpus, so we can find similar documents.  A dense vector representation is convenient for many applications, such as clustering of documents. \citet{DBLP:conf/icml/LeM14} extended word2vec with the paragraph vector to embed documents as well. This is called doc2vec in the popular implementation gensim\footnote{\texttt{http://radimrehurek.com/gensim/}}. \citet{W16-1609} examine parameter tunings for doc2vec, using an unweighted sum of the word vectors as a baseline.  

Recently \citet{arora2016asimple} looked at sentence embedding methods using ideas similar to the present work. They suggest using a weighted sum of word vectors as a basic embedding. They introducted a new word weight function called smooth inverse frequency (SIF), which is discussed further in Section \ref{sectionweights}. They also suggested a post process applied to the weighted sum they called common component removal. They apply Principal Component Analysis (PCA) to the document vectors $\mathbf{u}_d$ collected as a matrix, and find the first principal component $\mathbf{p}$. Then they subtract the projection of the vector to the first principal component:
\begin{equation}\label{pcaprojection}
\mathbf{u}_d = \mathbf{u}_d - \mathbf{p} (\mathbf{p} \cdot \mathbf{u}_d)
\end{equation}
We will experiment with these ideas in later sections. 

We derive two extensions of using a weighted sum for a document embedding, which center each vector by comparing the document term frequency to the corpus term frequency.  We experimentally compare these two forms to a weighted sum, using 3 possible weighting functions, both with and without the PCA post processing. These approaches do better than doc2vec on our data, and are competitive on the CQADupStack benchmark.

\section{An optimal embedding for a document}\label{sec:docembed}

Let $\mathbf{v}_i \in \mathbb{R}^K$ be a the vector associated with word $i \in V$ for some given vector space embedding of a text corpus with vocabulary $V$. Assume that these vectors have all been normalized to have a unit Euclidean norm, so $||\mathbf{v}_i|| = 1, \forall{i}$. 

Consider a single document within the corpus, with $n_i$ being the count of word $i$ in the document.  If $N = \sum_i{n_i}$ then the term frequencies for word $i$ in the document are $\text{tf}_i = n_i/N$.  We also have corpus wide words counts $n_{ic}$ with $N_c = \sum_i{n_{ic}}$, giving corpus wide term frequencies $\text{tf}_{ic} = n_{ic}/N_c$.

A weighted average of the word vectors in a document is perhaps the simplest type of document embedding:
\begin{equation}\label{simplesum}
\mathbf{c} = \sum_i{w_i \text{tf}_i \mathbf{v}_i}.
\end{equation}

The raw $\text{tf}_i$ in (\ref{simplesum}) is not that informative by itself.  Is a $\text{tf}_i = 0.001$ in a document high or not?  It is more useful to know if a word is over or under represented in a document relative to the corpus.  Let $\delta_i = \text{tf}_i - \text{tf}_{ic}$, so a positive $\delta_i$ means word $i$ is overrepresented in our document, and a negative $\delta_i$ means $i$ is underrepresented. You could think of $\delta_i$ as the net term frequency of word $i$ in the document, relative to the corpus.

Let $\mathbf{u} \in \mathbb{R}^K$ be the unknown vector embedding of our document within the same vector space, constrained to also have a unit norm, $||\mathbf{u}|| = 1$. The unit norm constraint is is convenient, because the cosine similarity we would like to maximize reduces to the dot product $\mathbf{v}_i \cdot \mathbf{u}$. 

We want to find the unit norm vector $\mathbf{u}$ that maximizes the total net cosine similarity $\delta_i (\mathbf{v}_i \cdot \mathbf{u})$ of the document with each word $i$, where word $i$ is weighted by a positive weight $w_i$:
\begin{equation}
\max_{\mathbf{u}}{\sum_i{w_i \delta_i \mathbf{v}_i \cdot \mathbf{u}}} \;\;  \text{subject to} \;\;  ||\mathbf{u}|| = 1.
\end{equation}
The $\delta_i$ term will cause $\mathbf{u}$ to move toward overrepresented words, and away from underrepresented ones.  

If we collect the coefficients for $\mathbf{u}$ into a vector $\mathbf{c}$:
\begin{equation}
\mathbf{c} = \sum_i{w_i \delta_i \mathbf{v}_i},
\end{equation}
we have
\begin{equation}\label{cleanprob}
\max_{\mathbf{u}}{\mathbf{c} \cdot \mathbf{u}} \;\; \text{subject to} \;\; ||\mathbf{u}|| = 1.
\end{equation}
The optimal solution of (\ref{cleanprob}) is the vector $\mathbf{u}^*$ that is pointing in the same direction as the objective vector $\mathbf{c}$ on the unit norm:
\begin{equation}\label{renormalize}
\mathbf{u}^* = \frac{\mathbf{c}}{||\mathbf{c}||}.
\end{equation}
While intuitively clear, a rigorous proof that (\ref{renormalize}) is the optimal solution to (\ref{cleanprob}) follows from the Cauchy-Schwarz inequality.

We can expand the $\delta_i$ in the unnormalized optimal solution $\mathbf{c}$:
\begin{equation}\label{expanded}
  \mathbf{c} = \sum_i{w_i \text{tf}_i \mathbf{v}_i} -  
  \sum_i{w_i \text{tf}_{ic} \mathbf{v}_i}.
\end{equation}
Let $V_d \subseteq V$ be the words $i$ that actually are in our document, and thus have $n_i > 0$ and a positive term frequency $\text{tf}_i$. Taking advantage of sparsity, the first sum of (\ref{expanded}) can be expressed with $V_d$:
\begin{equation}\label{expandedsparse}
\mathbf{c} = \sum_{i \in V_d}{w_i \text{tf}_i \mathbf{v}_i} - \sum_{i \in V}{w_i \text{tf}_{ic} \mathbf{v}_i}.
\end{equation}
Note that the second term of (\ref{expandedsparse}) is independent of the document, and could be viewed as a weighted center point of the corpus, so we will refer to (\ref{expandedsparse}) as the ``center'' form. The center point can be computed once and reused to efficiently embed multiple documents. 

In the center form, every word affects the embedding of every document through the center point. Intuitively, we would not expect the absence of a rare word to provide any information to an embedding.  We could choose a threshold $\text{tf}_{ic}$, and only include words above this threshold in the second term.  Another approach that avoids a tunable parameter (and worked better in practice) is to only include words $i$ in the second term that appear in our document.  That is, to also take the second sum over $V_d$:
\begin{equation}\label{deltaform}
\mathbf{c} = \sum_{i \in V_d}{w_i \text{tf}_i \mathbf{v}_i} - \sum_{i \in V_d} {w_i \text{tf}_{ic} \mathbf{v}_i}.
\end{equation}

We could rewrite (\ref{deltaform}) as:
\begin{equation}\label{compactdeltaform}
\mathbf{c} = \sum_{i \in V_d}{\delta_i \text{idf}_i \mathbf{v}_i},
\end{equation}
so we refer to (\ref{deltaform}) or (\ref{compactdeltaform}) as the ``delta'' form.  

In Section \ref{experimentssection} we will see that the center form works better on shorter documents, while the delta form is better on longer ones.

\section{Weights}\label{sectionweights}

The previous section did not specify a word weight $w_i$. The weight most used in Information Retrieval (IR) along with term frequency would be the inverse document frequency $\text{idf}_i = \log(D/D_i)$, where $D$ is the total number of documents, and $D_i$ is the number of documents containing word $i$. 

We can see that optimal solution for (\ref{expandedsparse}) with $w_i = \text{idf}_i$ is the tf-idf weight\-ed sum of the $\mathbf{v}_i$ using term frequency computed for the document, minus the tf-idf weighted sum of the $\mathbf{v}_i$ using term frequency computed over the corpus, then renormalized according to (\ref{renormalize}). 

\citet{arora2016asimple} recently proposed smooth inverse frequency (SIF) as a weight function: 

\begin{equation}\label{sif}
w_i = \frac{a}{a + \text{tf}_{ic}},
\end{equation}
where $a$ is a parameter we set to $10^{-4}$. They provide a theoretical justification for this weighting using a generative model for sentences. They also note that this function is mathematically similar to the function used by \citet{Mikolov:2013:DRW:2999792.2999959} to subsample frequent words\footnote{\citet{Mikolov:2013:DRW:2999792.2999959} was considering the probability to omit a word, so this is 1 minus their formula}:
\begin{equation}\label{mikolov}
w_i =
   \begin{cases}
     \sqrt{\frac{t}{\text{tf}_{ic}}},    & \text{if tf}_{ic} \ge t\\
    1.0,     & \text{if tf}_{ic} < t
   \end{cases}
\end{equation}
where $t$ is a parameter we set to $10^{-5}$.

\begin{figure}
\centering
\includegraphics[width=3in]{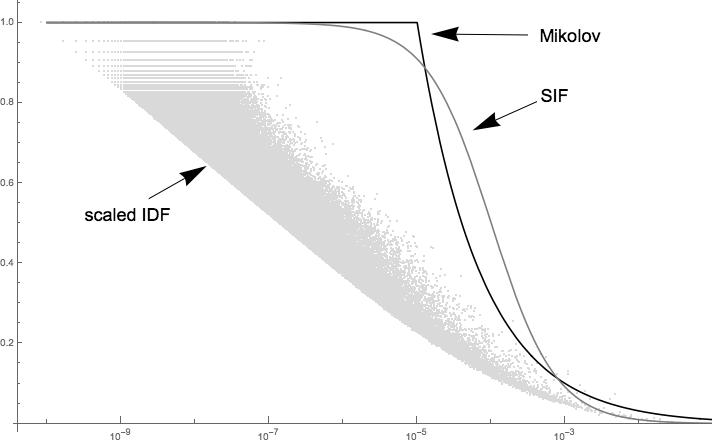}
\caption{Comparison of weight functions}\label{Fi:1}
\end{figure}

Figure \ref{Fi:1} shows the three weight functions, where the $\text{idf}_i$ used in Section \ref{experimentssection} have been scaled by
\begin{equation}
\frac{\text{idf}_i - \min_i{\text{idf}_i}}{\max_i{\text{idf}_i} - \min_i{\text{idf}_i}},
\end{equation}
so they line in the range of 0 to 1.  The equations for (\ref{sif}) and (\ref{mikolov}) are quite close to each other at our parameter settings. Both approach 1 more quickly than the idf values. 

We will consider all 3 weight functions in the benchmark in Section \ref{experimentssection}.  \citet{arora2016asimple} reported that SIF was better than idf on their test problems, but we found SIF and (\ref{mikolov}) to both be slightly but consistently worse than idf.
 
\section{TripAdvisor benchmark}

This section presents a new benchmark problem for document embedding, based on \mbox{TripAdvisor} reviews. TripAdvisor is the largest travel website, containing hundreds of millions of reviews of hotels, restaurants, and attractions.  The structure of the benchmark was inspired by \citet{hoogeveen2015}, discussed in the next section. 

Given a corpus containing multiple reviews for a number of locations, we would prefer an embedding that puts the reviews of a given location closer to each other than to the reviews of the other locations. We can use this idea to provide labels for a supervised benchmark. 

The dataset is constructed by selecting the 20 most reviewed hotels, 20 most reviewed attractions, and 20 most reviewed restaurants.   

For each location, we will create 250 documents from reviews.  Then we will compute the cosine similarity for each pair of the 15,000 documents. We would like the cosine similarity of documents in the same location to be above those for different locations.  We could view the cosine similarity as prediction scores in a classification problem, and label pairs in the same location as a 1, and pairs in different locations as a 0.  The area under the curve (AUC) of the receiver operating characteristic (ROC) is a metric that will measure the quality of the scores. It will be 1.0 if all of our pairs in the same location have scores above those in different locations, and will be 0.5 if they was no relation between the score and label.  

We would like to examine the effect of document lengths on the embeddings, so we create a series of 20 problems.  For problem $k \in 1,\dots,20$, a document will be the concatenation of $k$ randomly selected reviews. Thus, the $k=20$ dataset will use $20*250=5,\!000$ reviews for each location. 

Providing more data in longer documents makes the problem easier, so the ROC AUC increases with the number of reviews per document.  We will be interested in the relative order of the embeddings as the document length changes. 

We use a 100 dimensional word2vec embedding trained using the default settings of the original word2vec implementation, which was trained over all 9 billion words of English language \mbox{TripAdvisor} reviews.  We compute the $\text{tf}_{ic}$ and $\text{idf}_{i}$ over the full set of reviews as well. 

\section{TripAdvisor benchmark experiments}\label{experimentssection}

We would like to explore all the possible variations of our approach for document embedding.  We can use a simple weighted sum (\ref{simplesum}), the center form (\ref{expandedsparse}), or the delta form (\ref{deltaform}).  We have 3 possible weight functions: idf, SIF, and Mikolov's function (\ref{mikolov}).  Further we can do PCA common component removal with (\ref{pcaprojection}) as a post process or not. We compute these 18 variations over all 20 benchmark problems.  

\begin{figure*}[p]
\centering
\includegraphics[width=6.5in]{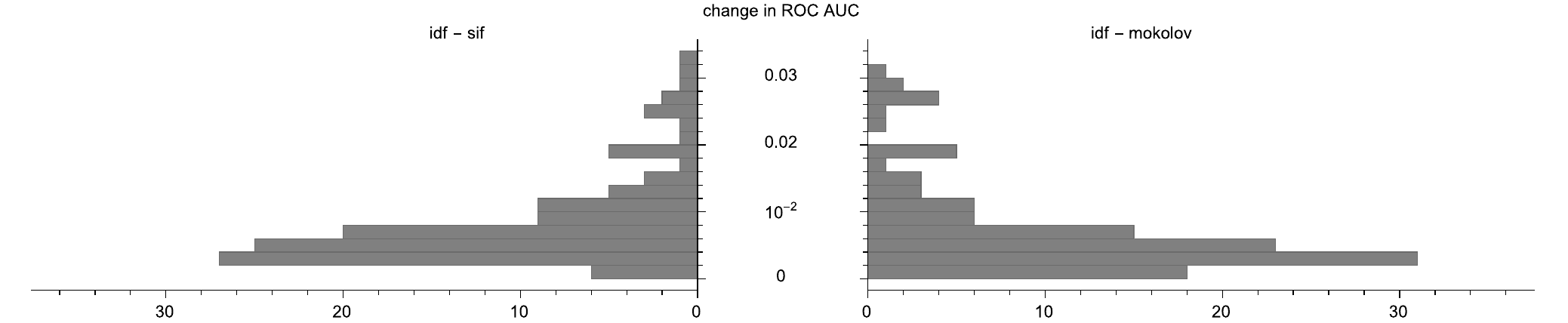}
\caption{Comparison of weight functions}\label{Fi:2}
\end{figure*}

The weight function turned out to be a dimension that had a dominant solution.  We can consider all $6 * 20$ problems for each weight, and then examine the change in ROC AUC between the idf weight and each of the other two weight functions.  All 120 of these paired comparisons were positive in both cases, indicating idf was always better.  Two histograms showing the change in ROC AUC are given in Figure \ref{Fi:2}. This is contrary to experience of \citet{arora2016asimple}, who found SIF superior.  This could be due to the large corpus used to compute the $\text{idf}_{i}$ in our problem, which is a superset of our problem's reviews. They were also working at a sentence level, while we had a mean document length of 103 words or longer.

The ROC AUC results for the six idf weighted variations of the 20 benchmark problems are given in Table \ref{Ta:taresults}. 

An obvious baseline approach is to take the unit weight sum of all the vectors for each word in the document, as was used in \citet{W16-1609}, which is the unit-sum column in Table \ref{Ta:taresults}.

We also trained a doc2vec model on the test documents using gensim \citep{rehurek_lrec}.  \citet{W16-1609} examined how to tune the parameters of doc2vec for several similar tasks.  They found the \texttt{dbow} approach superior to \texttt{dmpv} in most cases, so we use that approach. We create embeddings using their preferred settings\footnote{\texttt{size=300, window=15, min\_count=1 or min\_count=5,
sample=1e-5,  alpha=0.025, min\_alpha=0.001, dm=0, negative=5, dbow\_words=1, dm\_concat=1,  iter=20}}, although we do use a \texttt{min\_count} word count of both 1 and 5, since it seems that the optimal setting of \texttt{min\_count} will depend on document length. These are the columns d2v-mc1 and d2v-mc5 in Table \ref{Ta:taresults}, respectively.

\begin{table*}[p]
\centering\small
\begin{tabular}{rccccccccc}
  \toprule
$k$ & idf-sum & idf-sum-pca & idf-center & idf-center-pca & idf-delta & idf-delta-pca & unit-sum & d2v-mc1 & d2v-mc5 \\
  \midrule
1 & 0.7916 & \textbf{0.8185} & 0.8060 & 0.7713 & 0.7889 & 0.8182 & 0.6890 & 0.7091 & 0.7436 \\
2 & 0.8767 & \textbf{0.8999} & 0.8842 & 0.8622 & 0.8747 & 0.8998 & 0.7781 & 0.8309 & 0.8473 \\
3 & 0.9121 & 0.9301 & 0.9159 & 0.9056 & 0.9115 & \textbf{0.9304} & 0.8296 & 0.8788 & 0.8904 \\
4 & 0.9315 & 0.9455 & 0.9331 & 0.9300 & 0.9319 & \textbf{0.9461} & 0.8614 & 0.9043 & 0.9146 \\
5 & 0.9444 & 0.9555 & 0.9445 & 0.9454 & 0.9454 & \textbf{0.9563} & 0.8848 & 0.9224 & 0.9295 \\
6 & 0.9574 & 0.9620 & 0.9590 & 0.9553 & 0.9581 & \textbf{0.9629} & 0.9120 & 0.9344 & 0.9381 \\
7 & 0.9645 & 0.9673 & 0.9651 & 0.9628 & 0.9654 & \textbf{0.9683} & 0.9243 & 0.9432 & 0.9438 \\
8 & 0.9694 & 0.9715 & 0.9695 & 0.9685 & 0.9704 & \textbf{0.9725} & 0.9323 & 0.9497 & 0.9483 \\
9 & 0.9731 & 0.9748 & 0.9729 & 0.9726 & 0.9742 & \textbf{0.9758} & 0.9394 & 0.9536 & 0.9510 \\
10 & 0.9764 & 0.9775 & 0.9759 & 0.9761 & 0.9774 & \textbf{0.9786} & 0.9451 & 0.9576 & 0.9533 \\
11 & 0.9789 & 0.9798 & 0.9784 & 0.9789 & 0.9800 & \textbf{0.9808} & 0.9497 & 0.9600 & 0.9549 \\
12 & 0.9814 & 0.9820 & 0.9807 & 0.9814 & 0.9825 & \textbf{0.9830} & 0.9533 & 0.9618 & 0.9568 \\
13 & 0.9831 & 0.9836 & 0.9823 & 0.9832 & 0.9842 & \textbf{0.9846} & 0.9567 & 0.9634 & 0.9580 \\
14 & 0.9846 & 0.9852 & 0.9839 & 0.9850 & 0.9857 & \textbf{0.9861} & 0.9595 & 0.9651 & 0.9602 \\
15 & 0.9860 & 0.9866 & 0.9852 & 0.9865 & 0.9870 & \textbf{0.9875} & 0.9621 & 0.9663 & 0.9611 \\
16 & 0.9870 & 0.9876 & 0.9863 & 0.9875 & 0.9880 & \textbf{0.9885} & 0.9640 & 0.9677 & 0.9618 \\
17 & 0.9882 & 0.9886 & 0.9874 & 0.9887 & 0.9892 & \textbf{0.9895} & 0.9660 & 0.9685 & 0.9627 \\
18 & 0.9891 & 0.9896 & 0.9883 & 0.9896 & 0.9901 & \textbf{0.9904} & 0.9673 & 0.9697 & 0.9638 \\
19 & 0.9900 & 0.9904 & 0.9893 & 0.9904 & 0.9910 & \textbf{0.9912} & 0.9690 & 0.9707 & 0.9647 \\
20 & 0.9907 & 0.9910 & 0.9899 & 0.9910 & 0.9916 & \textbf{0.9918} & 0.9707 & 0.9710 & 0.9653 \\
\bottomrule
\end{tabular}
  \caption{ROC AUC for TripAdvisor benchmark}\label{Ta:taresults}
\end{table*}

\begin{table*}[p]
\centering\small
\begin{tabular}{rcccccccc}
  \toprule
group & idf-sum & idf-sum-pca & idf-center & idf-center-pca & idf-delta & idf-delta-pca & unit-sum & doc2vec \\
  \midrule
android & 0.917 & 0.984 & 0.974 & 0.982 & 0.928 & \textbf{0.985} & 0.768 & 0.956 \\
english & 0.672 & \textbf{0.832} & 0.755 & 0.797 & 0.678 & 0.831 & 0.619 & 0.816 \\
gaming & 0.929 & 0.991 & 0.984 & 0.990 & 0.936 & \textbf{0.992} & 0.879 & 0.947 \\
gis & 0.861 & 0.939 & 0.926 & \textbf{0.949} & 0.871 & 0.938 & 0.786 & 0.827 \\
mathematica & 0.711 & 0.883 & 0.804 & 0.845 & 0.727 & \textbf{0.886} & 0.653 & 0.846 \\
physics & 0.954 & 0.957 & 0.936 & \textbf{0.968} & 0.956 & 0.960 & 0.808 & 0.908 \\
programmers & 0.856 & 0.954 & 0.939 & 0.948 & 0.874 & \textbf{0.954} & 0.749 & 0.906 \\
stats & 0.730 & 0.898 & 0.878 & 0.880 & 0.747 & 0.902 & 0.702 & \textbf{0.924} \\
tex & 0.784 & 0.891 & 0.850 & 0.890 & 0.793 & 0.890 & 0.750 & \textbf{0.903} \\
unix & 0.797 & 0.915 & 0.873 & 0.901 & 0.813 & 0.924 & 0.776 & \textbf{0.949} \\
webmasters & 0.839 & 0.926 & 0.901 & 0.916 & 0.851 & \textbf{0.930} & 0.771 & 0.877 \\
wordpress & 0.845 & 0.930 & 0.888 & 0.874 & 0.869 & 0.933 & 0.613 & \textbf{0.980} \\
\bottomrule
\end{tabular}
  \caption{ROC AUC for CQA benchmark}\label{Ta:stackresults}
\end{table*}

The highest value in each row of Table \ref{Ta:taresults} is shown in bold. The idf-sum-pca form (use an idf weighted sum with a PCA post process) is best for 1 and 2 reviews per document. Then the idf-delta-pca form is best for 3 to 20 reviews per document.

Considering the non-PCA cases, we can see that idf-center and idf-delta cross between 4 and 5 reviews, with idf-sum sandwiched in between. From Table \ref{Ta:sizes}, an average document length around 450 words is the crossover point with idf-center better on shorter documents. 

The doc2vec models with \texttt{min\_count=1} and \texttt{min\_count=5} also cross between 7 and 8 reviews per document, or at about 775 words, with \texttt{min\_count=5} better on shorter documents. In this benchmark, the doc2vec models were not competitive with the idf weighted models. 

The PCA post process always helps for the sum and delta cases, but results are mixed for center.  With the exception of the 5 review case, Non-PCA is better for 1 to 9 reviews, and PCA is better for 10 reviews and up. It would appear that for shorter documents the center form is already doing a sufficient job of removing the common components.

The unit-sum baseline trailed far behind all other cases, indicating that at the very least a simple tf-idf weighted combination of the word vectors should used instead.

\section{StackExchange benchmark}\label{stackexchange}

\citet{hoogeveen2015} created a benchmark dataset called CQADupStack, using questions from 12 StackExchange question answering forums. 
One task involves finding questions that have been marked as duplicates.  
The different forums have between 1.52\% and 9.31\% duplicates. 
The questions vary in average length from 83.4 words for the \texttt{english} forum to 166.5 words for \texttt{programmers}. 
They specify a large subset of pairs of questions to use in scoring cosine similarity.  
Pairs with a duplicate question are labeled as 1, and other pairs a labeled as 0. 
They use the scores and labels to compute the ROC AUC for each forum, inspiring the approach in the previous section.

\citet{W16-1609} use this dataset in their evaluation of doc2vec. 
In one section they compare a doc2vec model and skip-gram word2vec models trained on a Wikipedia dump and then applied to CQADupStack. 
Rather than doing up to hundreds of millions of pairs in each forum, they randomly selected 10 million pairs for each forum.   
We recreate the \texttt{dbow WIKI} and \texttt{skip-gram WIKI} results in their Table 5, which correspond to our doc2vec and unit-sum columns, respectively. We add the six idf weighted variations we used in Section \ref{experimentssection}, with results given in Table \ref{Ta:stackresults}.  In this case $\text{idf}_i$ are computed over the set of forum data.

Results were more mixed on this benchmark.  The doc2vec model was best in 4 forums, idf-delta-pca was best in 5 forum, and idf-center-pca was best in 2 forums. The forum with the shortest text, \texttt{english}, was best with idf-sum-pca.  Once again, a PCA post process was always helpful for sum and delta, but mixed for center. The unit weighted sum trailed distantly behind in all cases. 

\section{Conclusions}

We give a new benchmark problem for document embedding based on TripAdvisor reviews. It prefers embeddings that put reviews for the same location closer than those of other locations, which is an idea that could be adapted to other benchmark problems. 

We use this benchmark to examine a number of variations of approaches for document embedding.  We find that at shorter lengths, the idf-sum-pca method with the PCA post process suggest by \citet{arora2016asimple} is best. Perhaps due to our good estimates for $\text{idf}_i$, the SIF weight function they suggested lagged slightly behind. At longer lengths of 300 words or more, the idf-delta-pca approach using (\ref{deltaform}) performed the best.  

If we want an extremely easy to implement solution without a PCA post process, then the idf-center form becomes best for short lengths below 450 words, and idf-delta is best for longer lengths. 

Clearly, document length needs considered when designing vector embeddings, and no single approach may be best for all lengths. 



\bibliography{acl2017}
\bibliographystyle{acl_natbib}

\appendix

\pagebreak

\section{Supplemental Material}\label{sec:supplemental}

\begin{table}[htp]
\centering\small
\begin{tabular}{rlll}
  \toprule
$k$ & min words & mean words & max words \\
  \midrule
1 & 9 & 103.6 & 4771 \\
2 & 30 & 207.3 & 4825 \\
3 & 53 & 311.6 & 4926 \\
4 & 81 & 415.5 & 5411 \\
5 & 107 & 516.8 & 6106 \\
6 & 136 & 620.1 & 6822 \\
7 & 164 & 722.0 & 6490 \\
8 & 219 & 824.3 & 6939 \\
9 & 227 & 927.8 & 7644 \\
10 & 259 & 1031.6 & 8302 \\
11 & 297 & 1133.9 & 8180 \\
12 & 315 & 1235.8 & 9015 \\
13 & 374 & 1338.4 & 8605 \\
14 & 371 & 1440.2 & 8862 \\
15 & 422 & 1542.5 & 10014 \\
16 & 407 & 1646.6 & 9455 \\
17 & 475 & 1749.9 & 9921 \\
18 & 462 & 1853.8 & 10577 \\
19 & 530 & 1956.8 & 10950 \\
20 & 551 & 2059.6 & 11263 \\\bottomrule
\end{tabular}
  \caption{Document lengths for the TripAdvisor benchmarks}\label{Ta:sizes}
\end{table}

\end{document}